\documentclass[a4paper,twoside]{article}

\usepackage{epsfig}
\usepackage{subcaption}
\usepackage{calc}
\usepackage{amssymb}
\usepackage{amstext}
\usepackage{amsmath}
\usepackage{amsthm}
\usepackage{multicol}
\usepackage{pslatex}
\usepackage{apalike}
\usepackage{tabularx}
\usepackage{color}
\usepackage{multirow}
\usepackage{multicol}
\usepackage{hyperref}

\hypersetup{
    colorlinks=true,
    linkcolor=magenta,
    filecolor=black,      
    urlcolor=black,
    citecolor=black
}

\newcolumntype{L}[1]{>{\raggedright\arraybackslash}p{#1}}
\newcolumntype{C}[1]{>{\centering\arraybackslash}p{#1}}
\newcolumntype{R}[1]{>{\raggedleft\arraybackslash}p{#1}}

\usepackage{SCITEPRESS}     
\graphicspath{ {fig/} }
\newcommand{\ch}[1]{\textcolor{black}{#1}} 
\def\eg{\emph{e.g. }} 
\def\ie{\emph{i.e. }}
\newcommand{\etal}{\textit{et al.}}

\begin{document}

\title{Normalized Convolution Upsampling  \\ for Refined Optical Flow Estimation}

\author{\authorname{Abdelrahman Eldesokey \sup{1}\orcidAuthor{0000-0003-3292-7153}, and Michael Felsberg \sup{1}\orcidAuthor{0000-0002-6096-3648}}
\affiliation{\sup{1}Computer Vision Laboratory, Link\"oping University, Sweden }
\email{\{abdelrahman.eldesokey@liu.se, michael.felsberg@liu.se\}}}


\abstract{Optical flow is a regression task where convolutional neural networks (CNNs) have led to major breakthroughs.
However, this comes at major computational demands due to the use of cost-volumes and pyramidal representations.
This was mitigated by producing flow predictions at quarter the resolution, which are upsampled using bilinear interpolation during test time.
Consequently, fine details are usually lost and post-processing is needed to restore them.
\ch{We propose the Normalized Convolution UPsampler (NCUP), an efficient joint upsampling approach to produce the full-resolution flow during the training of optical flow CNNs.
Our proposed approach formulates the upsampling task as a sparse problem and employs the normalized convolutional neural networks to  solve it.
We evaluate our upsampler against existing joint upsampling approaches when trained end-to-end with a a coarse-to-fine optical flow CNN (PWCNet) and we show that it outperforms all other approaches on the FlyingChairs dataset while having at least one order fewer parameters.
Moreover, we test our upsampler with a recurrent optical flow CNN (RAFT) and we achieve state-of-the-art results on Sintel benchmark with $\sim 6\%$ error reduction, and on-par on the KITTI dataset, while having 7.5\% fewer parameters (see Figure \ref{fig:1}).
Finally, our upsampler shows better generalization capabilities than RAFT when trained and evaluated on different datasets.}
}

\onecolumn \maketitle \normalsize \setcounter{footnote}{0} \vfill

\section{\uppercase{Introduction}}

Computer vision encompasses a broad range of regression tasks where the goal is to produce numerical output given a visual input. 
Some of these tasks such as depth prediction and optical flow even require pixel-wise output, which makes theses tasks more challenging.
Convolutional neural networks (CNNs) have lead to major breakthroughs in these regression tasks by exploiting deep representations of data.
A common design for these regression CNNs is coarse-to-fine where a low-resolution prediction is produced and then progressively upsampled and refined to the full-resolution.
This usually requires \ch{abundant} GPU memory, especially at finer stages as the spatial dimensionality grows. 
Therefore, the scale of these networks has been throttled by the availability of computational resources, which has been mostly mitigated either by limiting the depth of the networks or reducing the resolution of the data.

\setlength{\tabcolsep}{1pt}
\renewcommand{\arraystretch}{0.8}
\begin{figure}[]
	\begin{tabular}{cc}
		RAFT  & \textbf{RAFT+NCUP (Ours)} \\
		\includegraphics[width=0.235\textwidth, trim={0 0cm 0 0cm},clip]{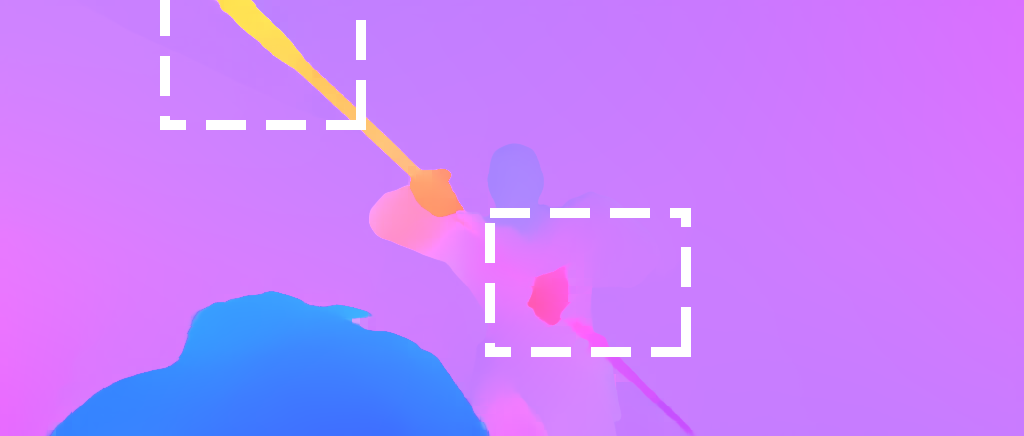} & \includegraphics[width=0.235\textwidth, trim={0 0cm 0 0cm},clip]{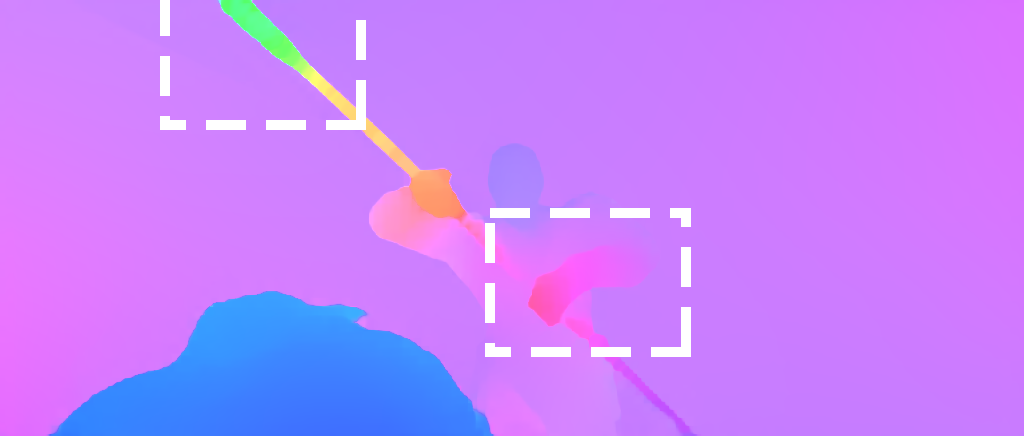} \\
		\includegraphics[width=0.235\textwidth, trim={0 0cm 0 0cm},clip]{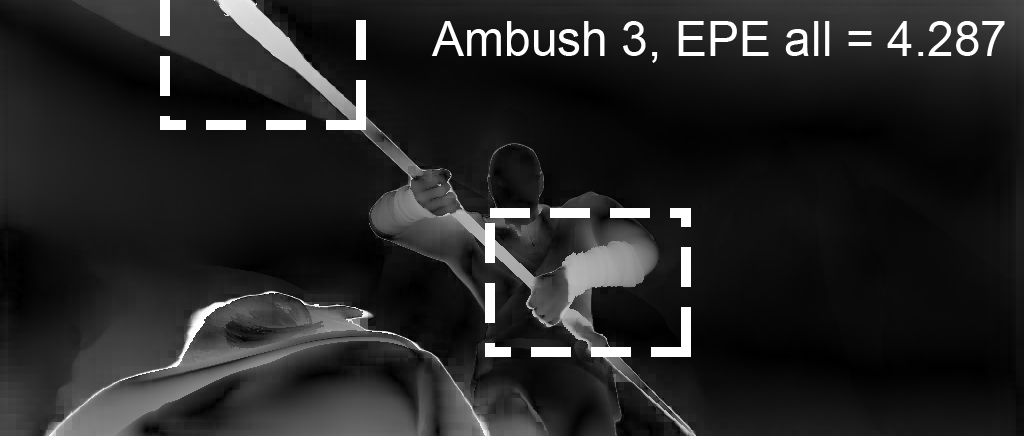} & \includegraphics[width=0.235\textwidth, trim={0 0cm 0 0cm},clip]{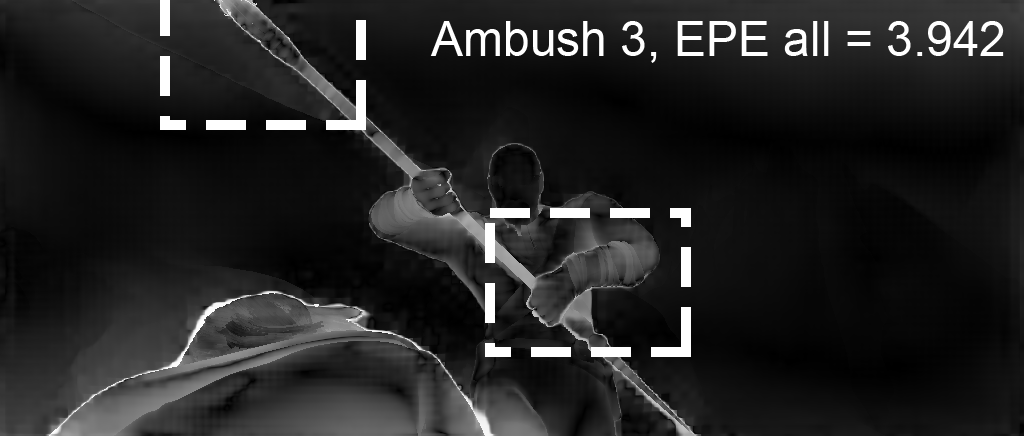} \\
	\end{tabular}
	\caption{An example from the Sintel \cite{sintel} test set that shows the flow improvement achieved by our proposed upsampler NCUP in comparison with RAFT \cite{raft}. }
	\label{fig:1}
\end{figure}

As an example, the early work on CNN-based depth estimation in \cite{eigen} employed an encoder/decoder network where the training datasets were downsampled to half the resolution to fit into the available GPU memory.
Similarly, the prevalent optical flow estimation network, FlowNet \cite{flownet}, trains on a quarter of the full resolution and uses bilinear interpolation to restore the full-resolution during test time. 
This practice has been preserved in subsequent optical flow CNNs, particularly with the increased complexity of these networks and the emergence of the computationally expensive cost-volumes and pyramidal representations \cite{flownet,pwcnet,flownet2}. 
Nonetheless, pyramid levels with full and half the resolution were not utilized as they would not fit on the available GPU memory.
Unfortunately, operating on a fraction of the full-resolution leads to loss of fine details, which might be crucial in certain tasks.

\begin{figure*}[t]
    
    \begin{subfigure}[t]{0.48\textwidth}
        {\includegraphics[width=\columnwidth]{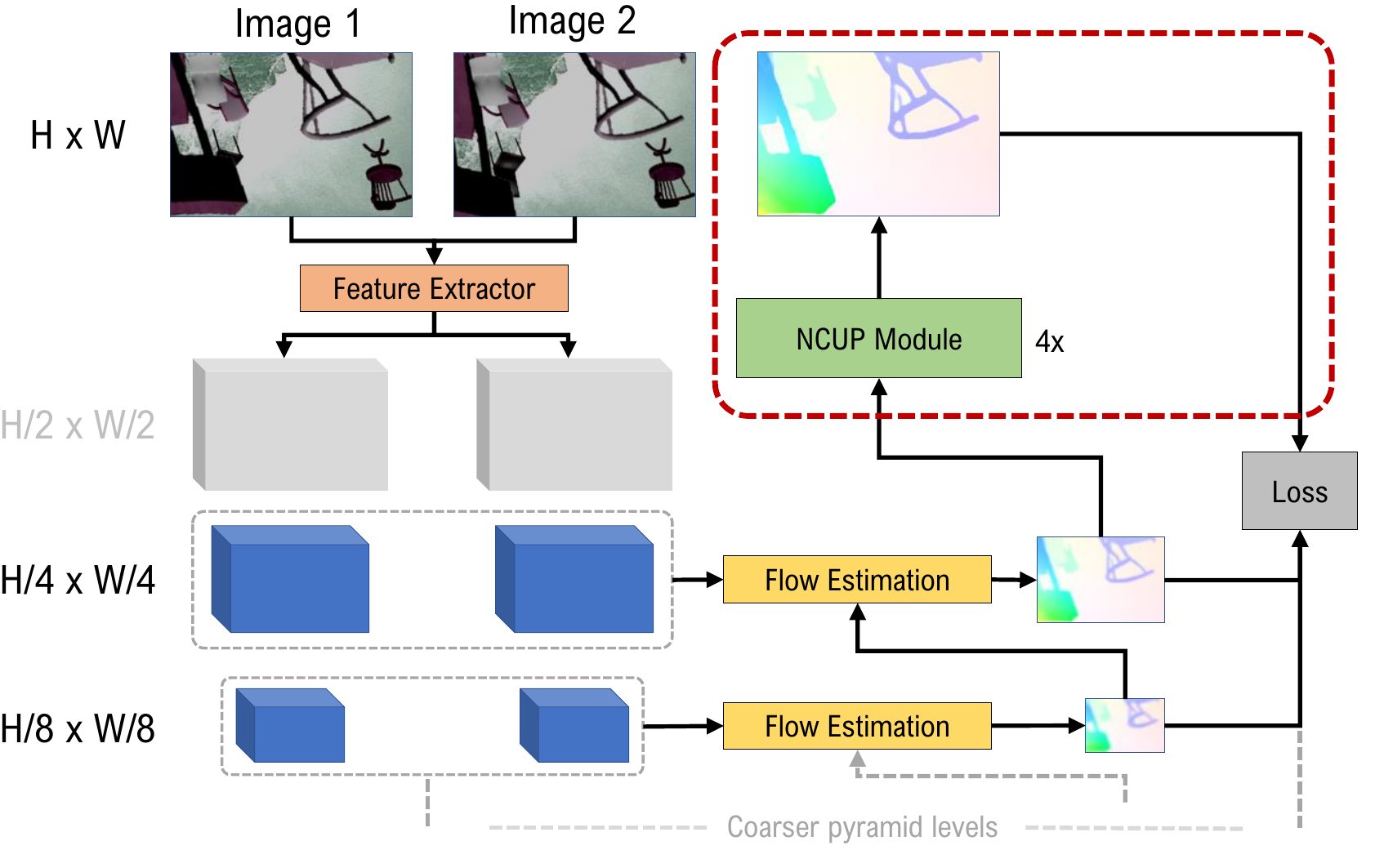}}
        \caption{Coarse-to-fine}
        \label{fig:intro_a}
    \end{subfigure}%
    ~
    \begin{subfigure}[t]{0.48\textwidth}
        \includegraphics[width=\columnwidth]{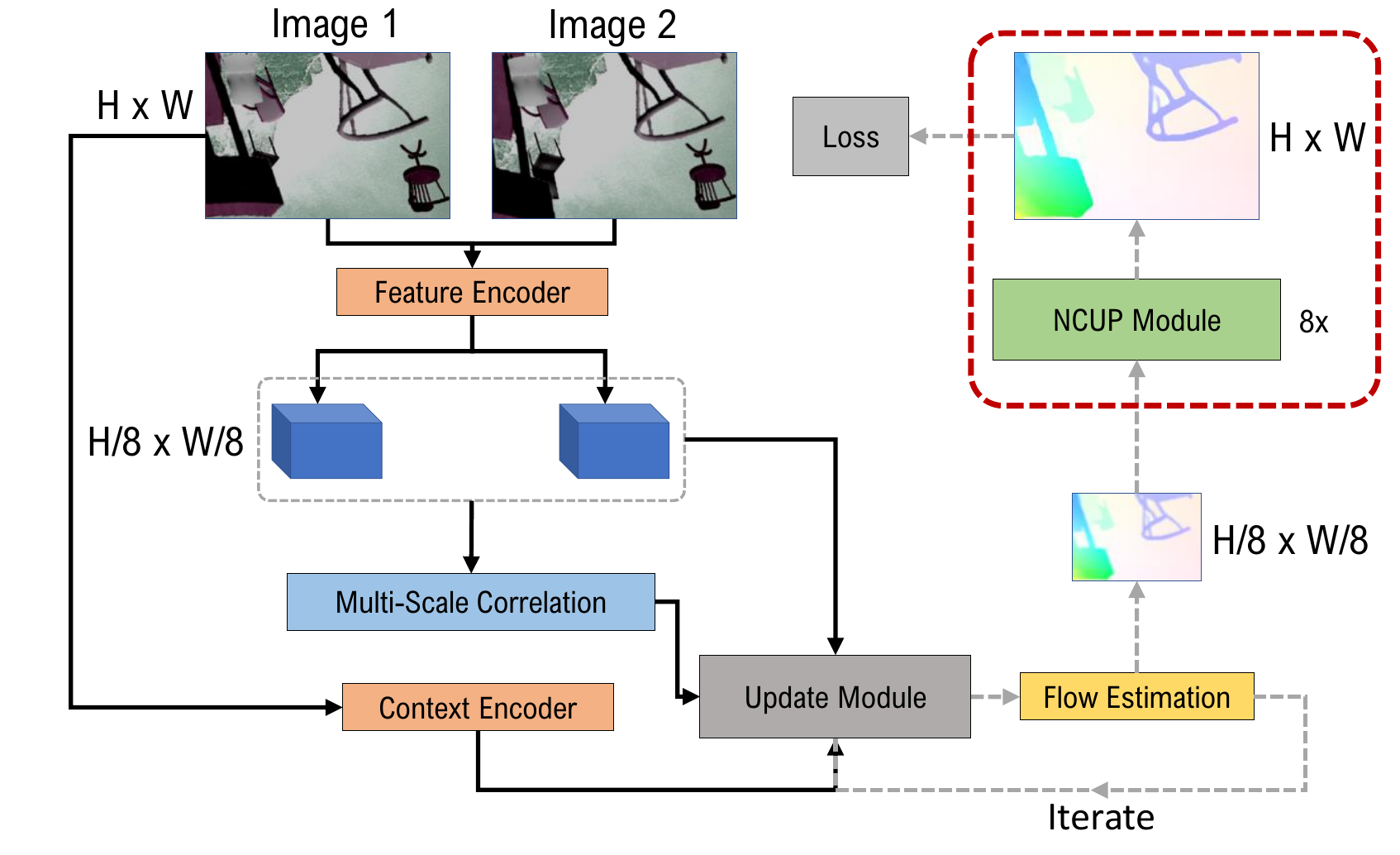}
        \caption{Recurrent}
        \label{fig:intro_b}
    \end{subfigure}
    \caption{An illustration for how we train our proposed normalized convolution upsampler (NCUP) with coarse-to-fine and recurrent optical flow networks.
    In coarse-to-fine CNNs, \eg PWCNet \cite{pwcnet} in \textbf{(a)}, the flow is estimated at different levels of a pyramid of features. 
    However, pyramid levels with full and half the resolution are not utilized as it is not feasible to fit them in GPU memory. 
    We upsample the flow to the full-resolution during training using our proposed approach leading to refined flow predictions. 
    In recurrent CNNs, \eg RAFT \cite{raft} in \textbf{(b)}, the full-resolution flow needs to be available after each iteration.
    We replace the convex combination upsampler in RAFT, with our more compact upsampler NCUP and we achieve state-of-the-art results using fewer parameters.
    }
    \label{fig:intro}
\end{figure*}

To alleviate these shortcomings of \emph{coarse-to-fine} approaches, several joint image upsampling approaches have been applied as post-processing to the output from optical flow and depth estimation networks \cite{djif,pac,wu2018fast}.
These approaches substitute the bilinear interpolation and they utilize RGB images as guidance to perform adaptive upsampling for the predicted flow that preserves edges and fine details. 
The key idea of theses approaches is to use a guidance modality, \eg RGB images, to guide the upsampling of a target modality such as flow fields or depth values.
However, these approaches act as post-processing and are trained separately from the network of the original task, omitting potential benefits from training them end-to-end. 
Therefore, we \ch{investigate} training these joint upsampling approaches within the coarse-to-fine optical flow CNNs, \eg FlowNet, PWCNet, in an end-to-end fashion to allow optical flow networks to exploit the fine details during training.
Moreover, we propose a novel joint upsampling approach (NCUP) \ch{that formulates the upsampling as a sparse problem and employs the normalized convolutional neural networks \cite{bmvc,pami} to solve it.} 
Our proposed upsampler that is more efficient (2k parameters) and outperforms other joint upsampling approaches in comparison on the task of end-to-end optical flow upsampling.
An illustration for the proposed setup is shown in Figure \ref{fig:intro_a}.

Another category of optical flow networks that emerged recently is based on \emph{recurrent networks} \cite{irr,raft}, where the predicted flow is iteratively refined.
This requires the availability of the flow in full-resolution at the end of each iteration.
The bilinear interpolation was used for this purpose in \cite{irr}, while a learnable convex combination upsampler was used in \cite{raft}.
However, this convex upsampler performs the upsampling with a scaling factor of 8 in a single-shot with a limited kernel support of $3 \times 3$.
Moreover, it has a large number of parameters which encompasses approximately 10\% of the entire network.
We replace this convex combination module with our efficient upsampler that performs the upsampling at multi-scales, leading to state-of-the-art results on Sintel dataset \cite{sintel}, \ch{similar results} on the KITTI dataset \cite{kitti}, better generalization capabilities, and using 5 times fewer parameters.
Figure \ref{fig:intro_b} shows an illustration for setup of recurrent networks, where we replace the upsampling module with our proposed upsampler.

\paragraph{Our Contributions can be summarized as follows:}
\begin{itemize}
    \item \ch{We propose a joint upsampling approach (NCUP) that formulates upsampling as a sparse problem and employs the normalized convolution neural networks to solve it.}
    \item We test our approach with \emph{coarse-to-fine} optical flow networks (PWCNet) to produce the full-resolution flow during training, and we show that it outperforms all other upsampling approaches, while having at least one order fewer parameters.
    \item When we use our upsampler with a recurrent optical flow CNN, \eg RAFT \cite{raft}, we achieve state-of-the-art results on the Sintel \cite{sintel} benchmark, and \ch{perform similarly} on the KITTI \cite{kitti} test set \ch{using 5 times less parameters than their convex combination upsampler.}
    \item We show that our upsampler has better generalization capabilities than the convex combination in RAFT, when trained on FlyingThings3D \cite{things} and evaluated on Sintel and KITTI.
\end{itemize}

\section{\uppercase{Related Work}}
\paragraph{CNN-based Optical Flow}
Deep learning recently surfaced as a plausible substitute for the classical optimization-based optical flow approaches \cite{xu2017accurate,bailer2015flow,horn1981determining}. 
CNNs can be trained to directly predict optical flow given two images avoiding explicitly designing an optimization objective manually in classical approaches.
FlowNet \cite{flownet} introduced the first CNN for optical flow estimation that is trained end-to-end in a coarse-to-fine fashion. 
Subsequent approaches followed the same scheme where FlowNet2 \cite{flownet2} proposed a stacked version of FlowNet, PWCNet \cite{pwcnet} introduced a pyramidal variation, and LiteFlowNet \cite{liteflownet} designed a light-weight cascaded network at each pyramid level.
VCN \cite{vcn} proposed several improvements for matching cost-volumes to expand their receptive field and they added support for multi-dimensional similarities.

Recently, several recurrent approaches were proposed where the flow is iteratively refined similar to the optimization-based approaches. 
An initial flow prediction is produced at the first iteration and it is refined for a number of iterations.
IRR \cite{irr} proposed to use either FlowNetS \cite{flownet} or PWCNet \cite{pwcnet} as a recurrent unit that iteratively estimates the residual flow from the previous iteration.
However, the number of iterations was limited either by the size of the network in FlowNet, or the number of pyramid levels in PWCNet.
RAFT \cite{raft} introduced a light-weight recurrent unit that is coupled with a GRU cell \cite{gru} as an update operator.
This cell allowed performing more iterations and led to refined flow predictions at a relatively lower computations.


\paragraph{Joint Image Upsampling}
The notion of joint (guided) image upsampling is to use a guidance image to steer the upsampling of another target image, where both the guidance and the target images could be from the same or different modalities.
Several classical approaches were proposed that are based on variations of the bilateral filtering \cite{yang2007spatial, barron2016fast}.
Li~\etal~\cite{djif} proposed a CNN-based architecture for joint image filtering that can be applied to joint upsampling. 
They employed two sub-networks for target and guidance features extraction followed by a fusion block.
Wu~\etal~\cite{wu2018fast} proposed a trainable guided filtering network that was applied to clone the behavior of several vision tasks. 
Su~\etal~\cite{pac} proposed pixel-adaptive convolutions that modifies the convolution filter with a spatially varying kernel.
Wannenwetsch \etal \cite{ppac} extended the pixel-adaptive convolutions to incorporate pixel-wise confidences.

\paragraph{Optical Flow Upsampling}
For coarse-to-fine networks, FlowNet \cite{flownet} suggested the use of an iterative variational approach \cite{brox2010large} to produce the full-resolution flow during test time.
However, this approach is computationally expensive and is not possible to train jointly with the network.
For recurrent networks, the full-resolution flow is required during the training at the end of each iteration.
IRR \cite{irr} attempted a residual upsampling block, but found to be futile with optical flow and they used the bilinear interpolation.
RAFT \cite{raft} produces the flow in 1/8 of the full-resolution and employed a convex combination upsampler to construct the full-resolution.
However, their upsampler has a limited receptive field and has a large number of parameters.

For coarse-to-fine networks, \ch{we look into employing differentiable joint upsampling approaches to upsample the flow during training.
Moreover, we propose a joint upsampling approach (NCUP) that maps the upsampling task to a sparsity densificiation problem and employ the efficient normalized convolutional neural networks \cite{bmvc,pami} to solve it.
Experiments show that our upsampler performs better than other approaches in comparison on optical flow upsampling.}
Different to other joint upsampling approaches, our upsampler estimates the guidance on the low-resolution data instead of the full-resolution ones, \ch{which leads to fewer computations and memory requirements compared to other approaches.}

For recurrent networks, \ie RAFT \cite{raft}, we replace the convex module with our proposed upsampler, which performs the upsampling at multi-scales and has 5 times fewer parameters.
\ch{This modification leads to state-of-the-art results on Sintel \cite{sintel} dataset with $\sim 6\%$ error reduction, similar performance on the KITTI \cite{kitti} dataset, while using 7.5\% fewer parameters.}
Finally, our approach shows better generalization capabilities when trained on FlyingThings \cite{things} and tested on Sintel and the KITTI datasets.

\section{\uppercase{Approach}}

\begin{figure*}[t]
    \centering
    {\includegraphics[width=\textwidth]{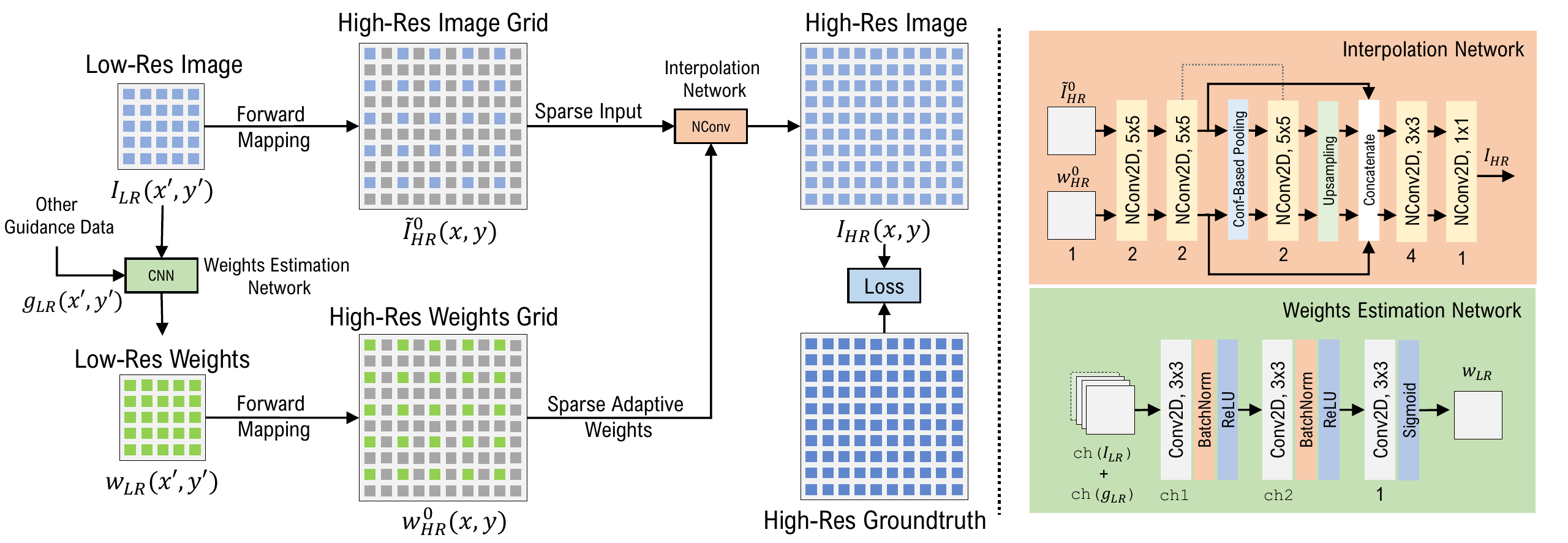}}
    \caption{An illustration of our proposed joint upsampling approach (NCUP). First, \ch{a sparse high-resolution grid is constructed from the low-resolution image using forward mapping}. Pixel-wise weights for the low-resolution image are produced by a \emph{weights estimation network the} (green block) which takes the low-resolution image and other auxiliary data as input. The weights are mapped to the high-resolution grid in a similar fashion using forward mapping. Next, an \emph{interpolation network} that encompasses a cascade of normalized convolution layers (the orange block) receives the high-resolution grid as well as the weights and produce the high-resolution image. Note that the notation \texttt{ch()} denotes number of channels.}
    \label{fig:method}
\end{figure*}

In joint image upsampling task, it is desired to train a network $\mathbf{\theta}$ to upsample a low-resolution input $\mathbf{I}_\text{LR}$ to a high-resolution output $\mathbf{I}_\text{HR}$, guided by some high-resolution guidance data $\mathbf{g}_\text{HR}$; $\mathbf{\theta}: \mathbf{I}_\text{LR} \rightarrow \mathbf{I}_\text{HR} | \mathbf{g}_\text{HR}$.
The guidance data is typically the RGB image, but can be of any modality or even intermediate feature representations from a CNN.
In this section, we briefly describe the normalized convolutional neural networks \cite{bmvc} followed by our proposed Normalized Convolution Upsampler (NCUP).


\subsection{Normalized Convolutional Neural Networks}
Eldesokey \etal \cite{bmvc, pami} proposed the normalized convolution layer, which is a sparsity-aware convolution operator that was used to interpolate a sparse depth map on an irregular grid.
More formally, they learn an interpolation function $\mathbf{\theta}: \mathbf{\tilde{I}}_\text{HR} \rightarrow \mathbf{I}_\text{HR} | \tau(\mathbf{\tilde{I}}_\text{HR})$, where $\mathbf{\tilde{I}}_\text{HR}$ is a sparse high-resolution input with missing pixels, and $\tau()$ is a thresholding operator that produces ones at pixels where data is present and zeros otherwise.
They recently proposed to replace the thresholding operator $\tau$ with a CNN $ \Phi $ that predicts pixel-wise weights from the sparse input $\mathbf{\theta}: \mathbf{\tilde{I}}_\text{HR} \rightarrow \mathbf{I}_\text{HR} | \Phi(\mathbf{\tilde{I}}_\text{HR})$ in a self-supervised manner \cite{cvpr}.
The high-resolution output $\mathbf{I}_\text{HR}$ is predicted by a cascade of $L$ normalized convolution layers, where the output for layer $l 
\in \{1 \dots L \}$, is calculated as:
\begin{equation}
\mathbf{I}_\text{HR}^l(\mathbf{x}) = \dfrac{\sum_{\mathbf{m} \in \mathbb{R}^2} \ \mathbf{I}_\text{HR}^{l-1}(\mathbf{x}-\mathbf{m}) \ \mathbf{w}^{l-1}(\mathbf{x}-\mathbf{m}) \ \mathbf{a}^l(\mathbf{m}) }{\sum_{\mathbf{m} \in \mathbb{R}^2} \ \mathbf{w}^{l-1}(\mathbf{x}-\mathbf{m}) \ \mathbf{a}^l(\mathbf{m})} \enspace,    
\end{equation}
where $\mathbf{x}, \mathbf{m}$ are the spatial coordinates of the image, $\mathbf{I}_\text{HR}^0=\mathbf{\tilde{I}}_\text{HR}$, $\mathbf{w}^0=\Phi(\mathbf{\tilde{I}}_\text{HR})$, and $\mathbf{a}^l$ is the interpolation kernel at layer $l$. The weights are propagated between layers as:
\begin{equation}
\mathbf{w}^l(\mathbf{x}) = \dfrac{ \sum_{\mathbf{m} \in \mathbb{R}^2} \ \mathbf{w}^{l-1}(\mathbf{x}-\mathbf{m}) \ \mathbf{a}^l(\mathbf{m})} {\sum_{\mathbf{m} \in \mathbb{R}^2} \mathbf{a}^l(\mathbf{m})} \enspace,    
\end{equation}
At the final layer $L$, the high-resolution output is produced $\mathbf{I}_\text{HR} = \mathbf{I}_\text{HR}^L$.


\subsection{Formulating Upsampling as a Sparse Problem}
Typically, the standard interpolation operations, \eg bilinear, bicubic, employ backward mapping to ensure that each pixel in the output is assigned a value.
Contrarily, if forward mapping is used, a sparse grid is formed in the output.
Fortunately, normalized convolution layers were demonstrated to perform well with irregular sparse grids, \eg depth completion, sparse optical flow, and, consequently, can be used to interpolate regular sparse grids.

Given a low-resolution input image $\mathbf{I}_\text{LR}$, a high-resolution sparse grid $\mathbf{\tilde{I}}_\text{HR}$ can be constructed using forward mapping.
The forward mapping from the low-resolution grid coordinates $(x',y')$ to the high-resolution grid $(x,y)$ for a scaling factor $s$ can be realized as:
\begin{equation}
(x,y) = (\text{round}(s \cdot x' ), \text{round}(s \cdot y') ) \ \forall \ (x',y')
\end{equation}
Note that the high-resolution grid is regular when $s \in \mathbb{N} $.

The initial pixel-wise weights $\mathbf{w}^0$ required for the normalized convolution network can be estimated using a weights estimation network $\Phi$ similar to \cite{cvpr}.
But different from \cite{cvpr} and other existing joint upsampling approaches, we estimate the pixel-wise weights from the low-resolution guidance image, not the high-resolution one.
\ch{Predicting weights for the low-resolution image requires less computations and memory requirements  making the weights estimation network much smaller and shallower, and therefore, leading to a more efficient upsampling.}
For instance, the entire upsampling network that we use with coarse-to-fine optical flow networks, \eg FlowNet and PWCNet, has only 2k parameters (see Figure \ref{fig:method} where \texttt{ch1}=16 and \texttt{ch2}=8), while being able to outperform other approaches with at least one order of magnitude more parameters.

Another difference from \cite{cvpr} is that we employ other modalities, \eg RGB input image, intermediate CNN features, as guidance for the weights estimation network similar to the existing joint upsampling approaches \cite{djif,pac}; $\Phi([ \mathbf{I}_\text{LR}, \mathbf{g}_\text{LR}])$.
This allows exploiting other modalities to adapt the weights based on the context.
The output from the weights estimation network is also transformed to the high-resolution grid using the forward mapping.

Essentially, we train an upsampling network $\mathbf{\theta}: \mathbf{I}_\text{LR} \rightarrow \mathbf{I}_\text{HR} | \Phi([ \mathbf{I}_\text{LR}, \mathbf{g}_\text{LR}])$, where the sparse high-resolution grid $\mathbf{\tilde{I}}_\text{HR}$ is an intermediate stage generated by applying forward mapping to the the low-resolution input. 
The pixel-wise weights are predicted using a CNN from the low-resolution input and any other guidance data.
The weights are similarly mapped to the high-resolution grid using forward mapping.
Finally, a cascade of normalized convolution layers is applied to interpolate the missing values in the sparse high-resolution grid.
An illustration of the whole pipeline is shown in Figure \ref{fig:method}.


\subsection{Weights Estimation Network}
Since the weights are estimated for the low-resolution input, the receptive field of the weights estimation network can be quite small.
Therefore, we use two convolution layers with a $3 \times 3 $ filters followed by Batch Normalization and ReLU activation.
The number of channels per layer is determined based on the guidance data that is used.
When RGB images are used, we use 16 and 8 channels for the two convolution layers, while we use 64 and 32 channels when intermediate CNN features are used as guidance (\texttt{ch1} and \texttt{ch2} values in Figure \ref{fig:method}).
A last convolution layer with a $ 1 \times 1$ filter is applied to produce the same number of channels as the low-resolution input $\mathbf{I}_\text{LR}$.
Finally, a Sigmoid activation is applied to produce valid non-negative weights.
Other function with a non-negative co-domain can be used, \eg Softplus, but the Sigmoid function was found to achieve the best results.
The estimated weights are transformed to the high-resolution grid using forward mapping as well.


\subsection{Interpolation Network}
We build a U-Net shaped normalized convolution network inspired by \cite{bmvc}. 
However, we perform downsampling only once, \ie we use two scales instead of four in \cite{bmvc}, since the sparsity in our case is significantly lower than the LiDAR depth completion problem they were solving.
This leads to a smaller network with 224 parameters instead of 480 parameters in \cite{bmvc}.
The interpolation network receives the high-resolution image grid $\mathbf{\tilde{I}}_\text{HR}$ and the weights grid $\mathbf{w}^0$ as an input.
The weights are propagated and updated within the interpolation network until the final dense output $\mathbf{I}_\text{HR}$ is produced at the final layer.


\subsection{Optical Flow Upsampling}
Optical flow is represented as two channels for vertical and horizontal flow field.
We process the two channels jointly within the weights estimation network, \ie \texttt{ch}$(I_\text{LR}) =2$ in Figure \ref{fig:method}.
However, for the interpolation network, the two channels are processed separately and then concatenated.
In coarse-to-fine optical flow estimation networks, \eg FlowNet \cite{flownet} and PWCNet \cite{pwcnet}, the flow is produced at quarter the resolution.
We attach the upsampling module to the optical flow estimation network to upsample the flow from $H/4 \times W/4$ to $H\times W$

\renewcommand{\arraystretch}{1.2}
\setlength{\tabcolsep}{6pt}
\begin{table*}[t]
	\begin{center}
		\begin{tabular}{|L{2.8cm} || C{1.5cm} | C{1.5cm} | C{1.5cm} | C{1.5cm} | C{1.7cm} | C{1.7cm} |}
		    \hline
			 & Baseline & Bilinear & DJIF & PAC & ConvComb & \textbf{NCUP (Ours)} \\
			\hline
			\hline
            PWCNet \cite{pwcnet} & 1.69 & 1.58 (+6.5\%) & 1.51 (+10.6\%) & \textit{1.50} (+11.2\%) & 1.52 (+10.0\%) & \textbf{1.46} (+13.6\%) \\			
			\hline
			FlowNetS \cite{flownet} & 2.53 & 2.23 (+11.8\%) & 2.16 (+14.6\%) & \textbf{2.11} (+18.8\%) & 2.16 (+14.6\%) & \textit{2.13} (+15.8\%) \\
			\hline
			\hline
			Relative  Params. & - & - & +56k & +183k & \textit{+44k} & \textbf{+2k}\\
			\hline
		\end{tabular}
	\end{center}
	\caption{Summary of the results fpr two \emph{coarse-to-fine} optical flow networks trained end-to-end with joint upsampling approaches. Relative Params. indicates the number of parameters for each upsampler. The stated results are the Average End-Point Error (AEPE) on the FlyingChairs \cite{flownet} test set. The relative improvement is shown between parentheses. The best results are shown in \textbf{Bold} and the second best in \textit{Italics}. \ch{Our upsampler NCUP outperforms all other approaches DJIF \cite{djif}, PAC \cite{pac}, ConvComb \cite{raft} with PWCNet , while having the least number of parameters.}}
	\label{tab:c2f}
\end{table*}

Typically, the multi-scale loss is employed in coarse-to-fine networks:
\begin{equation}\label{eq:loss}
    \sum_{p \in P} \alpha_p \ | \mathbf{f}^p - \mathbf{f}^p_\text{GT} |^2 \enspace,
\end{equation}
where $\mathbf{f}^p$ is the flow estimation at pyramid level $p$ in PWCNet or resolution $p$ in FlowNet, where $P=\{3,4,5,6,7\}$,  and $\mathbf{f}^p_\text{GT} $ is the corresponding downsampled groundtruth. 
The choice of $\alpha_p$'s were empirically determined in \cite{flownet} as $\{ 0.32, 0.08, 0.02, 0.01, 0.005 \} $.
Note that that $p=1, p=2$, where not considered during training as explained earlier.
We consider another level/scale in the loss for the full-resolution flow, \ie we set $P=\{1,3,4,5,6,7\}$, and following \cite{flownet}, we found empirically that the best performance is obtained when $\alpha_1=0.02$ for most methods.
This denotes that the flow is upsampled by a factor of 4 from quarter the resolution to the full-resolution.
For the recurrent network RAFT, we use their proposed loss \cite{raft}.


\section{\uppercase{Experiments}}

\newcommand{\chexample}[2]{\includegraphics[width=0.24\textwidth, trim={0 0cm 0 0cm},clip]{qual-ch/#1_#2.png}
}
\setlength{\tabcolsep}{1pt}

\begin{figure*}[]
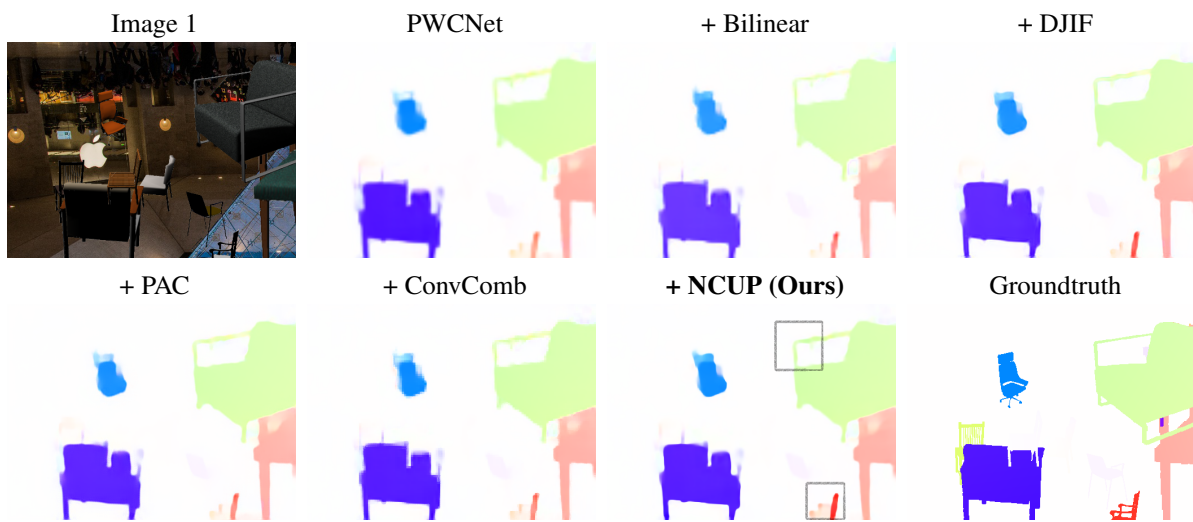

	\begin{tabular}{cccc}
		Image 1 & PWCNet & + Bilinear & + DJIF  \\
		\chexample{529}{img1} & \chexample{529}{pwc} & \chexample{529}{bi} & \chexample{529}{djif} \\ 
		+ PAC  & + ConvComb & \textbf{+ NCUP (Ours)} & Groundtruth \\
		\chexample{529}{pac} &  \chexample{529}{raft} & \chexample{529}{nc} & \chexample{529}{gt} \\
	\end{tabular}
	\caption{A qualitative example from the FlyingChairs \cite{flownet} dataset when PWCNet \cite{pwcnet} is trained end-to-end using different joint upsampling approaches. Our upsampler produces sharp edges and preserves fine details such as the arm of the green chair and the small red chair at the bottom. Better viewed on a computer display.}
	\label{fig:qual_ch}
\end{figure*}

In this section, we evaluate our proposed joint upsampling approach with \emph{two types} of optical flow estimation CNNs: coarse-to-fine and recurrent networks.


\subsection{Joint Upsampling for Coarse-to-fine Networks} 
We choose two of the most popular coarse-to-fine optical flow CNNs, \ie FlowNet \cite{flownet} and PWCNet \cite{pwcnet}.
Different joint upsampling approaches are attached to the two networks and they are trained end-to-end as illustrated in Figure \ref{fig:intro_a}.
The joint upsampling approaches that we compare against are DJIF \cite{djif}, PAC \cite{pac}, the convex combination from RAFT \cite{raft} which we refer to as ConvComb, and the bilinear interpolation.
We train only on  the FlyingChairs \cite{flownet} as its spatial resolution is smaller than its counterparts allowing training memory-demanding joint upsampling approaches.
For instance, PWCNet trained with PAC fully occupy a 32 GB V100 GPU when trained of FlyingChairs with a batch size of 3.
We use the official PyTorch implementations provided by the corresponding authors.


\noindent \textbf{Experimental Setup} 
We initialize FlowNetS and PWCNet using pretrained models on the FlyingChairs dataset, while the joint upsampling approaches are initialized randomly.
We train each network for 60 epochs with an initial learning rate of 0.0001 that is halved at epochs $\{20, 30, 40, 50, 55\}$.
Since we can only fit a batch size of 3 for PAC on a 32GB V100 GPU, we use a batch size of 4 for all other approaches for a fair comparison.
We use data augmentation as described in \cite{irr}.


\noindent \textbf{Quantitative Results}
Table \ref{tab:c2f} summarizes the results for coarse-to-fine networks.
All upsampling approaches lead to performance gains demonstrating the advantage from making the full-resolution flow available for coarse-to-fine networks during training.
On PWCNet, our upsampler achieves the best improvement over the baseline despite having at least one order of magnitude lower parameters than its counterparts, while other approaches performs comparably well.
On FlowNetS, our upsampler performs second best with a small margin to PAC.
We believe that the larger model of PAC allows it to refine the poor predictions from FlowNetS slightly better than our upsampler.


\noindent \textbf{Qualitative Results}
A qualitative example for different approaches on the FlyingChairs dataset is shown if Figure \ref{fig:qual_ch}.
All upsampling approaches make edges and details more sharp and defined compared to the standard PWCNet as a result of making the full-resolution flow available during training.
Nonetheless, PAC and our upsampler tend to produce sharpest results amongst all.
However, our upsampler does a better job preserving small objects in some situations such as the red chair at the bottom of the scene.


\subsection{Joint Upsampling for Recurrent Networks} 
\renewcommand{\arraystretch}{1.2}
\setlength{\tabcolsep}{7pt}
\begin{table*}[!t]
	\begin{center}
		\begin{tabular}{|C{2.2cm} || L{2.7cm} | C{0.8cm}  C{0.8cm} | C{0.8cm}  C{1.0cm} | C{0.8cm}   C{0.8cm} | C{1.0cm}|}
		    \hline
			 \multirow{2}{*}{Training Dataset} &  \multirow{2}{*}{Method} & \multicolumn{2}{c|}{Sintel (Train)} & \multicolumn{2}{c|}{KITTI (Train)} & \multicolumn{2}{c|}{Sintel (Test)} & KITTI  \\
			 & & \textit{Clean} & \textit{Final} & \textit{AEPE} & \textit{Fl-All} & \textit{Clean} & \textit{Final} & (Test)\\
			\hline
			\hline
			\multirow{15}{*}{C+T} & PWCNet \cite{pwcnet} & 2.55 & 3.93 & 10.35 & 33.7 & - & - & - \\
			 & LiteFlowNet \cite{liteflownet} & 2.48 & 4.04 & 10.39 & 28.5 & - & - & - \\
 			 & VCN \cite{vcn} & 2.21 & 3.67 & 8.36 & 25.1 & - & - & - \\
 			 & MaskFlowNet \cite{maskflownet} & 2.25 & 3.61 & - & 23.1 & - & - & - \\ 
 			 & FlowNet2 \cite{flownet2} & 2.02 & 3.54 & 10.08 & 30.0 & 3.96 & 6.02 & - \\
 			 & RAFT-Small \cite{raft} & 2.21 & 3.35 & 7.51 & 26.9 & - & - & - \\
 			 & RAFT \cite{raft} & \textit{1.43} & \textbf{2.71} & \textit{5.04} & \textbf{17.4} & - & - & - \\
 			 & \textbf{RAFT+NCUP} & \textbf{1.41} & \textit{2.75} & \textbf{4.83} & \textbf{17.4} & - & - & - \\
 			 \hline
 			 \hline
  		     \multirow{9}{*}{C+T+S+K+H} & PWCNet+ \cite{pwcnet+} & (1.71) & (2.34) & (1.50) & (5.30) & 3.45 & 4.60 & 7.27 \\
  		     & VCN \cite{vcn} & (1.66) & (2.24) & (1.16) & (4.10) & 2.81 & 4.40 & 6.30 \\
  		     & MaskFlowNet \cite{maskflownet} & - & - & - & - & 2.52 & 4.17 & 6.10 \\
  		     & RAFT$^*$ \cite{raft} & (0.77) & (1.27) & (0.63) & (1.50) & \textbf{1.61} & \textit{2.86} & \textbf{5.10} \\
  		     & \textbf{RAFT+NCUP}$^*$ & (0.71) & (1.09) & (0.67) & (1.68) & \textit{1.66} & \textbf{2.69} & \textit{5.14} \\
			\hline
		\end{tabular}
	\end{center}
	\caption{ 
	Summary for quantitative results when using our upsampler NCUP with the \emph{recurrent} network RAFT \cite{raft}. The best results are shown in \textbf{Bold} and the second best in \textit{Italics}. Different datasets are indicated as following: FlyingChairs \cite{flownet} $\rightarrow$ C, FlyingThings3D \cite{things} $\rightarrow$ T, Sintel \cite{sintel} $\rightarrow$ S, KITTI-Flow 2015 \cite{kitti} $\rightarrow$ K, and HD1K \cite{hd1k} $\rightarrow$ H. \ch{Results between brackets are training set score and hence not comparable. Note that we did not use FlyingThings3D and HD1K during finetuning for Sintel. We outperform RAFT on the challenging final pass of Sintel and perform similarly on the test set of KITTI, while having 5 times smaller upsampler. $^*$ Indicates that warm starts \cite{raft} were used.}
	}
	\label{tab:rec}
\end{table*}
\newcommand{\stexample}[2]{\includegraphics[width=0.245\textwidth, trim={0 0cm 0 0cm},clip]{qual-st/sintel_#1_#2.png}
\includegraphics[width=0.245\textwidth, trim={0 0cm 0 0cm},clip]{qual-st/sintel_#1_#2_err.png}
}
\setlength{\tabcolsep}{1pt}
\renewcommand{\arraystretch}{0.8}
\begin{figure*}[t]
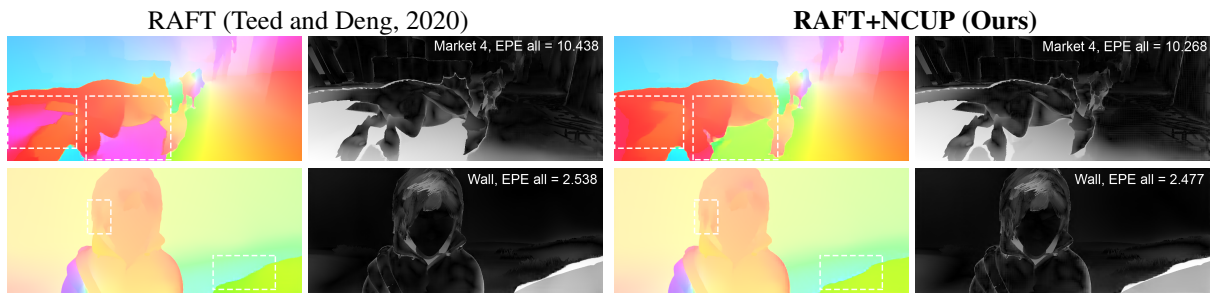

	\begin{tabular}{cc}
		RAFT \cite{raft} & \textbf{RAFT+NCUP (Ours)} \\
		\stexample{1}{raft} & \stexample{1}{nc} \\
		\stexample{4}{raft} & \stexample{4}{nc} \\
	\end{tabular}
	\caption{Qualitative examples from the Sintel \cite{sintel} test set.}
	\label{fig:qual_st}
\end{figure*}
We test our proposed upsampler as a substitute for the convex combination upsampler in the recurrent optical flow approach RAFT \cite{raft}.
The convex upsampler which has $\sim 500k$ parameters is removed and replaced with our upsampler constituting $\sim 100k$ parameters.
We use the output from the GRU cell, which has 128 channels as guidance data as they suggested in addition to the low-resolution flow.
For efficiency, we upsample the flow from 1/8 to 1/4 the full-resolution and then use our upsampler for restoring the full-resolution.

\noindent \textbf{Experimental Setup} 
We initialize the network using the pretrained weights provided by the authors \cite{raft}.
We use the same training hyperparameters as described in \cite{raft} except for the weight decay that we set to 0.00005 and we only train for 50k iterations.
For Sintel, we do not include FlyingThings3D and HD1k during finetuning.
For KITTI, we disable the batch normalization in the weights estimation network as it leads to better results.


\noindent \textbf{Benchmark Comparison}
Table \ref{tab:rec} shows the results for Sintel and KITTI benchmarks.
On the Sintel benchmark, we outperform the standard RAFT with a 6.3\% error reduction on the challenging final pass, while the error is slightly increased by 1.8\% on the clean pass.
We believe that this performance boost on the final pass is caused by multi-scale interpolation scheme employed by our upsampler that can eliminate large faulty regions in the predicted flow.
On the KITTI benchmark, we perform similarly the standard RAFT despite having 7.5\% fewer parameters.


\noindent \textbf{Generalization Results}
To examine the generalization capabilities of our upsampler, we train it on FlyingChairs followed by FlyingThings3D and evaluate it on the training set of Sintel and KITTI.
Table \ref{tab:rec} shows that our upsampler outperforms the standard RAFT on clean pass of Sintel and KITTI, while it performs slightly worse on the final pass of Sintel.
We believe that the slight degradation on the final pass is due to training the clean and the final pass of FlyingThings3D together without a weighted sampling.
However, the large improvement on KITTI significantly indicates that our upsampler posses better generalization.


\noindent \textbf{Qualitative Results}
Figure \ref{fig:qual_st} shows some qualitative results from the Sintel test set.
The use of our upsampler leads to better flow estimations compared to the standard RAFT.
The first row shows an example where a large region of faulty flow prediction (the purple region under the dragon) produced by the standard RAFT that is corrected when our proposed upsampler was used.
The second row shows another example where the flow is improved at fine details such as the hair.
These results clearly demonstrates the impact of upsampling on the quality of the flow.
{Qualitative examples for the KIITI dataset can be found on the online benchmark: \url{http://www.cvlibs.net/datasets/kitti/eval_scene_flow.php?benchmark=flow} }.


\subsection{Ablation Study}

\renewcommand{\arraystretch}{1.2}
\setlength{\tabcolsep}{7pt}
\begin{table}[t]
    \centering
    \begin{tabular}{|L{5.0cm}|C{2.0cm}|}
        \hline
         Model & AEPE  \\
         \hline
         \hline
         PWCNet+NCUP (Baseline) & \textbf{1.46}  \\
         \hline
         \multicolumn{2}{|c|}{\textit{Weights Estimation Network}} \\
         \hline
         Final activation is SoftPlus & 1.48 \\
         Estimate from High-Res & 1.75 \\
         Low-Res not used as guidance & 1.52 \\
         \hline
         \multicolumn{2}{|c|}{\textit{Interpolation Network}} \\
         \hline
         Two downsampling instead of one & 1.49\\ 
         Max instead of Conf. pooling & 1.48 \\
         \hline
         \multicolumn{2}{|c|}{\textit{Loss Function}} \\
         \hline
         $\alpha_1=0.002$ & 1.48 \\
         $\alpha_1=0.02$ & \textbf{1.46}  \\ 
         $\alpha_1=0.2$ & \textbf{1.46}  \\ 
         \hline
    \end{tabular}
    \caption{Ablation results on the FlyingChairs \cite{flownet} test set. The baseline is PWCNet \cite{pwcnet} trained with our upsampler NCUP.}
    \label{tab:abl}
\end{table}

We conduct an ablation study to justify specific design choices in our proposed approach.
Experiments are reported for PWCNet+NCUP on the FlyingChairs \cite{flownet} test set.
Table \ref{tab:abl} summarizes the average end-point-error scores for different experiments.

\noindent\textbf{Weights Estimation Network:} We replace the final activation with SoftPlus function instead of Sigmoid to get the estimate weights in the range of $[0, \infty[$ instead of $[0,1]$ produced by the Sigmoid function. 
The network converges faster when using the SoftPlus function, however the AEPE score is slightly worse.
We also attempt to feed the full-resolution guidance data to the weights estimation networks similar to other joint upsampling approaches.
The kernel size of the first two convolution layers was increased to $5 \times 5$ for a larger receptive field.
The results are significantly worse, which is probably because a larger network is needed to exploit the interesting information in the full-resolution data.
Finally, we omit the low-resolution flow from being used with guidance data.
The results shows that using the low-resolution flow with guidance data contributes significantly to the results.

\noindent\textbf{Interpolation Network:}
We experiment with two downsamplings, which indicates that the interpolation is performed at three scales instead of two.
The results show that the the best results are achieved when using only one downsampling.
We also test the standard max pooling for downsampling instead of the confidence-based pooling proposed in \cite{bmvc}.
The results show that the confidence-based pooling is slightly superior to max pooling.

\noindent\textbf{The Loss Function:} We experiment with one order of magnitude higher and lower factor $\alpha_1$ in
(\ref{eq:loss}). The results indicates that the choice of $\alpha_1=0.02$ and $\alpha_1=0.2$ lead to the best results. 
So, we choose $\alpha_1=0.02$ since it works the best for the majority of methods in comparison, but the value of $\alpha_1$ can be tuned further for our approach.


\subsection{What does our Upsampler Learn?}
\setlength{\tabcolsep}{1pt}
\renewcommand{\arraystretch}{0.8}
\begin{figure}[]
	\begin{tabular}{cc}
		\includegraphics[width=0.235\textwidth, trim={0 0cm 0 0cm},clip]{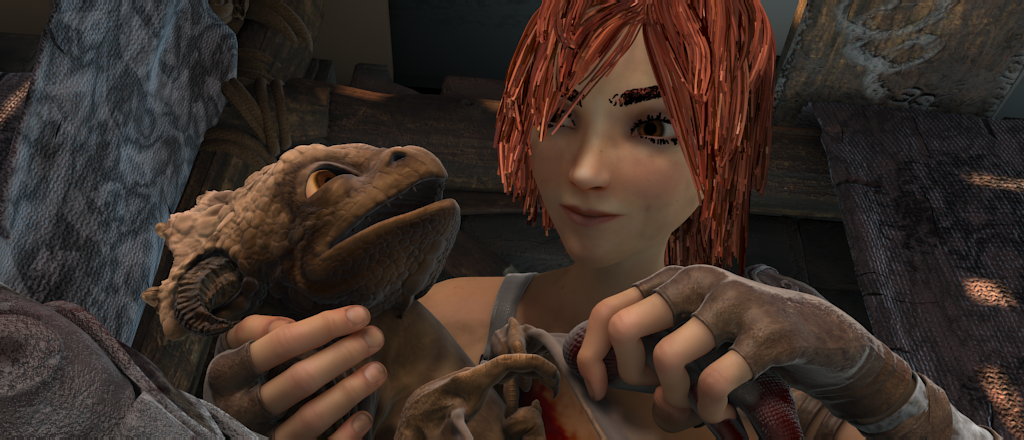} & \includegraphics[width=0.235\textwidth, trim={0 0cm 0 0cm},clip]{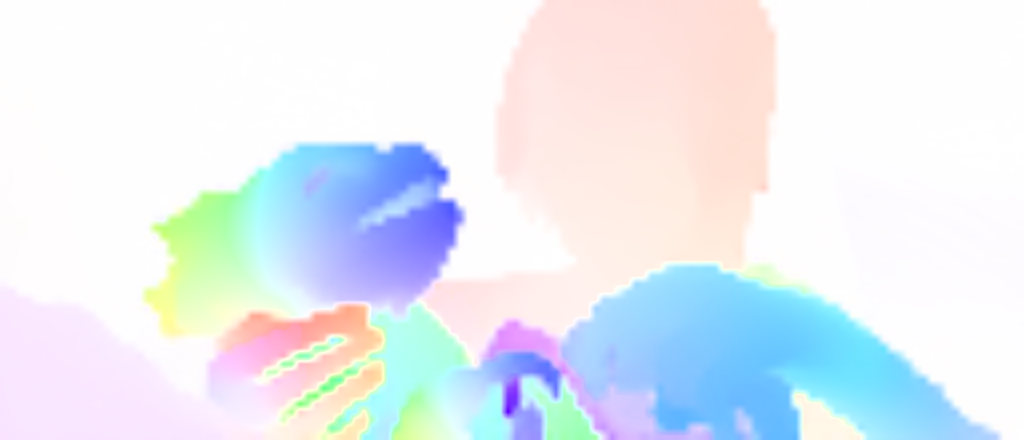} \\
		Image 1 & RAFT+Bilinear \\
		\includegraphics[width=0.235\textwidth, trim={0 0cm 0 0cm},clip]{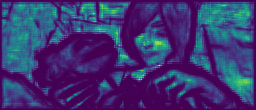} & \includegraphics[width=0.235\textwidth, trim={0 0cm 0 0cm},clip]{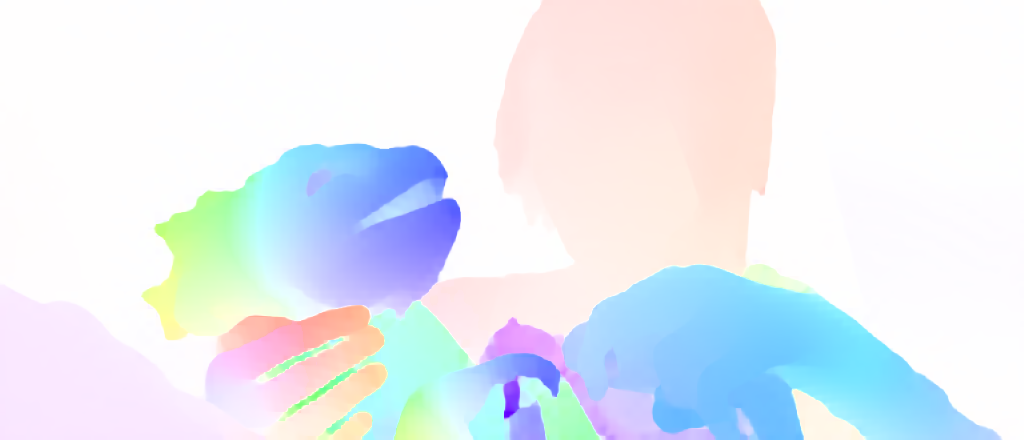} \\
		Estimated Weights & RAFT+NCUP \\
	\end{tabular}
	\caption{An example of the predicted weights from NCUP when used with RAFT \cite{raft}.}
	\label{fig:weights}
\end{figure}
Figure \ref{fig:weights} shows an example of the predicted weights within our upsampler when used with RAFT on the Sintel dataset in comparison to the bilinear interpolation.
The estimated weights essentially highlight edges and fine details with low-weight regions separating them.
The width of these regions defines to what extent each object is extrapolated and ensures the separability between objects.
Based on the design of the interpolation network, the width of these regions is adapted accordingly.
On the other hand, solid regions, \eg the girl's face, with no texture are assigned uniform weights acting as averaging.
This adaptive behavior shows a great potential for using our upsampling with other regression tasks, where the weights estimation network would learn the upsampling pattern that minimizes the reconstruction error.

\vspace{-5mm}
\section{\uppercase{Conclusion}}
We introduced an efficient upsampling approach based on the normalized convolutional networks that we incorporated in training coarse-to-fine and recurrent optical flow CNNs.
In coarse-to-fine networks, \eg PWCNet, the full-resolution flow was produced by our upsampler during the training leading to the fines flow estimations compared to other joint upsampling approaches in comparison, while having at least one order of magnitude fewer parameters.
When trained with the recurrent optical flow network RAFT, it achieved state-of-the-art results on the Sintel dataset, and achieved a similar score on the KITTI dataset, while having 400k less parameters.
Additionally, our approach showed better generalization capabilities compared to the standard RAFT.
\vspace{-2mm}
\section*{\uppercase{Acknowledgements}}
This work was supported by  the Wallenberg AI, Autonomous Systems and Software Program (WASP)
and Swedish Research Council grant 2018-04673. 

\bibliographystyle{apalike}
{\small
\bibliography{main}}

\end{document}